\documentclass{ecai} 

\usepackage{latexsym}
\usepackage{amssymb}
\usepackage{amsmath}
\usepackage{amsthm}
\usepackage{booktabs}
\usepackage{enumitem}
\usepackage{graphicx}
\usepackage{color}

\newcommand{\BibTeX}{B\kern-.05em{\sc i\kern-.025em b}\kern-.08em\TeX}

\begin{document}


\begin{frontmatter}


\paperid{846} 


\title{DiffSeg: A Segmentation Model for Skin Lesions Based on Diffusion Difference}


\author[A]{\fnms{Zhihao}~\snm{Shuai}}
\author[A]{\fnms{Yinan}~\snm{Chen}\footnotemark[1]}
\author[A]{\fnms{Shunqiang}~\snm{Mao}\footnotemark[1]}
\author[A]{\fnms{Yihan}~\snm{Zhou}\footnotemark[1]}
\author[A]{\fnms{Xiaohong}~\snm{Zhang}\thanks{Corresponding Author. Email: xhongz@cqu.edu.cn.}}

\address[A]{School of Big Data and Software Engineering, Chongqing University}


\begin{abstract}
Weakly supervised medical image segmentation (MIS) using generative models is crucial for clinical diagnosis. However, the accuracy of the segmentation results is often limited by insufficient supervision and the complex nature of medical imaging. Existing models also only provide a single outcome, which does not allow for the measurement of uncertainty. In this paper, we introduce DiffSeg, a segmentation model for skin lesions based on diffusion difference which exploits diffusion model principles to ex-tract noise-based features from images with diverse semantic information. By discerning difference between these noise features, the model identifies diseased areas. Moreover, its multi-output capability mimics doctors' annotation behavior, facilitating the visualization of segmentation result consistency and ambiguity. Additionally, it quantifies output uncertainty using Generalized Energy Distance (GED), aiding interpretability and decision-making for physicians. Finally, the model integrates outputs through the Dense Conditional Random Field (DenseCRF) algorithm to refine the segmentation boundaries by considering inter-pixel correlations, which improves the accuracy and optimizes the segmentation results. We demonstrate the effectiveness of DiffSeg on the ISIC 2018 Challenge dataset, outperforming state-of-the-art U-Net-based methods.
\end{abstract}

\end{frontmatter}


\begin{figure*}[ht]
    \centering
    \includegraphics[width=\textwidth]{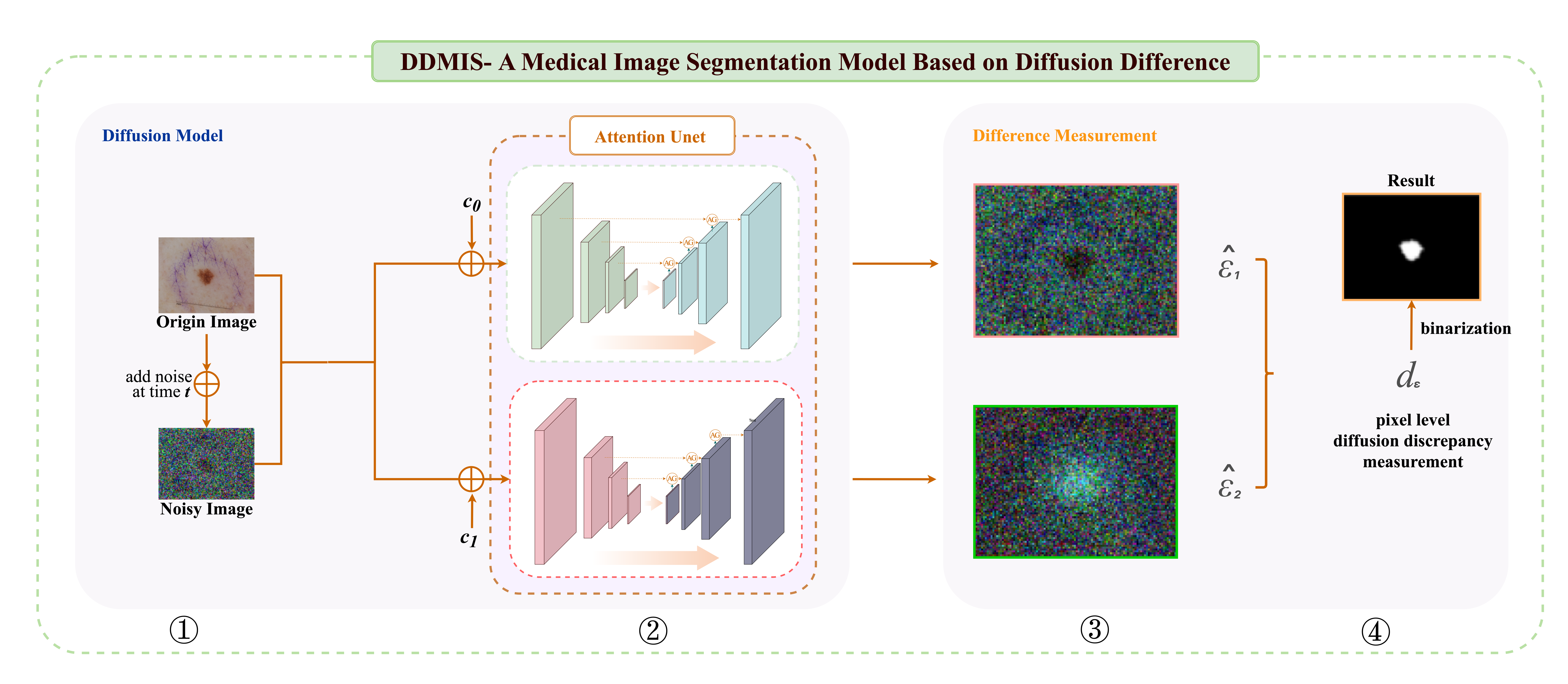} 
    \caption{The four steps of DiffSeg. First, add noise to the original image t times. Second, input the original image, noisy image, and category labels into the model. Third, guide the model to generate different noises through two category labels. Fourth, measure the difference between the two types of noise.}
    \label{fig:fig1_long}
\end{figure*}

\section{Introduction}

In the frontier exploration of medical image analysis, image segmentation technology is particularly crucial, serving as an essential foundation for achieving accurate diagnosis and effective treatment. As a foundational task in intelligent diagnostics, MIS is dedicated to precisely extracting key structures or regions from complex medical images, thereby providing robust support for doctors' disease diagnoses. 

In recent years, deep learning methods have received widespread attention in MIS tasks and have achieved significant accomplishments. These methods are largely based on Fully Convolutional Networks (FCN), particularly the U-Net model, which adopts an encoder-decoder structure. It effectively merges low-level and high-level information through skip connections, thereby significantly improving the accuracy of segmentation results. The advent of U-Net and its numerous variants has further propelled the development of MIS technology, enabling deep neural networks to process more complex and high-dimensional data, providing a powerful tool for medical image analysis. The advancement of these technologies not only enhances the performance of segmentation tasks but also offers more precise support for clinical diagnosis and treatment.

Existing MIS models often rely on a large amount of manually annotated data when dealing with complex images. Additionally, these models typically only provide a single output result, which has limited referential value. In light of this, the development of algorithms and technologies that can more effectively utilize unannotated or sparsely annotated data, as well as the design of optimization strategies for segmentation results under uncertainty \citep{abdarReviewUncertaintyQuantification2021}, has become a critical issue that urgently needs to be addressed in the field of MIS.

Considering the outstanding performance of diffusion models in the field of natural image processing, we propose a model for medical image segmentation that utilizes the differences in diffusion based on the Denoising Diffusion Probabilistic Models (DDPM). Although the input and output of diffusion models are strongly correlated with noise, their internal feature maps still possess discernible semantic information. Semantic segmentation can be achieved by extracting the semantic information hidden within the diffusion noise, which can be obtained by learning the differences between various noises.

The main contributions of this paper can be summarized as follows:

First, we develop an end-to-end framework from the original image to the optimized segmentation result based on diffusion difference. The model can extract hidden semantic information by learning the differences between various noises, achieving semantic segmentation under weak supervision. It is less susceptible to interference from other factors in the original image and does not depend on a complex annotation process.

Second, we propose a method for measuring the aleatoric uncertainty of segmentation results, which can simulate various segmentation solutions caused by unclear target structures and differences in doctors' experience. This method can intuitively display the consistent areas, ambiguous areas, and prediction confidence of the segmentation results and optimize the model's prediction results, thereby providing more assistance to doctors in practice.

Third, we define a segmentation result optimization method based on the multi-output capability of the diffusion model. This strategy employs DenseCRF algorithm to smooth out noise and local details. It aims to enhance the accuracy of segmentation and the interpretability of the model. This approach not only mitigates the issue of accidental uncertainty stemming from ambiguities in doctors' annotations but also contributes to improving the overall performance of the model.

The rest of the paper is organized as follows: In Section 2, we introduce the related work. In Section 3, we elaborate on our model architecture and optimization strategies. In Section 4, we first introduce the data preprocessing methods we adopted, followed by a description of the experimental setup and comparison methodologies, and finally present the experimental results. In Section 5, we compare our methods with previous studies and discuss the limitations of this paper as well as potential future research directions. Finally, we conclude the paper in Section 6.

\section{RELATED WORKS}
Research on MIS methods has a long history of development. Classic algorithms mainly include thresholding, edge detection, and region-based segmentation algorithms. Yet, these methods lack learning capabilities and highly depend on manually set parameters, quickly being surpassed by numerous methods in the deep learning field, such as fully convolutional networks \citep{longFullyConvolutionalNetworks2015}, \citep{otsuThresholdSelectionMethod1975}, \citep{sahooSurveyThresholdingTechniques1988a}, \citep{haralickImageSegmentationTechniques1985} and the U-Net proposed by Ronneberger et al. \citep{milletariVnetFullyConvolutional2016} However, since deep neural networks require a substantial amount of labeled data for training and given the complex semantics and cumbersome annotation process of medical images, the further development of deep learning algorithms in this field has been limited. 

Weakly supervised image segmentation algorithms, which reduce dependence on annotations, have received widespread attention. Currently, many scholars in this field have conducted extensive research using single models or collaborative training algorithms, achieving significant results. For example, Bai et al. \citep{baiSemisupervisedLearningNetworkbased2017a} combined conditional random fields' post-processing methods with self-training algorithms for ventricular MRI segmentation tasks. Similarly, Tang et al. \citep{tangDeepLevelSet2017} used the level set method for post-processing refinement of pseudo-labels. Additionally, Rajchl et al. \citep{rajchlDeepcutObjectSegmentation2016} used self-training methods with additional weak annotations at the boundary level to assist the supervision process. Zhou et al. \citep{zhouSemisupervised3DAbdominal2019} defined an additional student model based on collaborative training methods to learn fused pseudo-labels. Peng et al. \citep{pengDeepCotrainingSemisupervised2020} used the mean predictions of multiple models as pseudo-labels and introduced adversarial examples to capture differences between models.

In recent years, generative MIS technology has improved the advantages of strong segmentation modeling ability, good generalization, high efficiency, and low accuracy dependence on phosphate, and has become one of the core technologies to solve the problem of weakly supervised medical image segmentation. Jin et al. designed a Conditional Generative Adversarial Network (CGAN) \citep{jinCTRealisticLungNodule2018c} to learn the attribute dis-tribution of lung nodules in 3D space, improving the performance of the Progressive Holistically Nested Network (P-HNN) model for pathological lung segmentation in CT scans. Zhao et al. proposed a cascaded generative adversarial network with a deep supervision discriminator (Deep-supGAN) \citep{zhaoCraniomaxillofacialBonyStructures2018c} for skeletal segmentation, generating high-quality CT images from MRI for the supervisor to segment bone structures, enhancing segmentation accuracy. Dong et al. introduced a domain adaptation frame-work based on generative adversarial networks for measuring the cardiothoracic ratio (CTR) \citep{dongUnsupervisedDomainAdaptation2018}, incorporating discriminators to learn domain-invariant feature representations and generate accurate segmentations for unlabeled datasets. Yu et al. proposed a Shape Consistency Generative Adversarial Network (SC-GAN) \citep{yuAnnotationFreeCardiacVessel2019b} for coronary artery segmentation based on knowledge transfer, enhancing the accuracy of coronary artery segmentation. Liang et al. introduced a generative learning framework for semantic segmentation (GMMSeg) \citep{liangGMMSegGaussianMixture2022c}, establishing a Gaussian Mixture Model through expectation maximization to obtain class-conditional densities, showcasing the advantages of generative classifiers in large-scale visual tasks.

Simultaneously, diffusion models have emerged as a new generative model with powerful generation capabilities, achieving significant accomplishments, especially in the image domain. Guillaume, Jakob, and others proposed a semantic image editing method based on diffusion models \citep{couaironDiffEditDiffusionbasedSemantic2022c}, which automatically generates a mask by comparing the differences in the results of diffusion models under different text prompts. This method effectively annotates targets from images, providing new ideas and methods for the medical image segmentation field.

Furthermore, in the medical image segmentation field, due to subjective standards of doctors and objective cognitive differences among experts, a single image may have various segmentation solutions. These differences caused by insurmountable objective factors are referred to as Aleatoric Uncertainty \citep{wangAleatoricUncertaintyEstimation2019}. However, current work tends to output a single segmentation result, ignoring the uncertainty of the predictions. Doctors hope that the model can provide a quantification of the uncertainty of the results, so they can focus their efforts where the model's uncertainty is high, reducing repetitive work. Nowadays, different methods have been proposed to quantify the uncertainty in deep learning models, including evaluating uncertainty using Bayesian deep networks \citep{galDropoutBayesianApproximation2016b}, estimating arbitrary uncertainty through network output \citep{kendallWhatUncertaintiesWe2017}, and using deep ensembles for simple and scalable uncertainty prediction \citep{lakshminarayananSimpleScalablePredictive2017}, but the development has been somewhat limited.

Moreover, research on correcting segmentation results using the output uncertainty of models has also made some progress in fields like medical image segmentation. Yarin Gal et al. proposed Bayesian SegNet \citep{kendallMultitaskLearningUsing2018}, which can correct segmentation results through Bayesian inference on the model's output uncertainty. Ren et al. reduce the occurrence of false positives or false negatives in segmentation results by considering the uncertainty of model outputs \citep{pinheiroLearningRefineObject2016}. Zhang et al. optimized segmentation results by fusing segmentation outcomes from different modalities based on multi-modal data fusion segmentation results \citep{zhangModalityAwareMutualLearning2021a}. 

\begin{figure*}[t] 
    \centering
    \includegraphics[width=\textwidth]{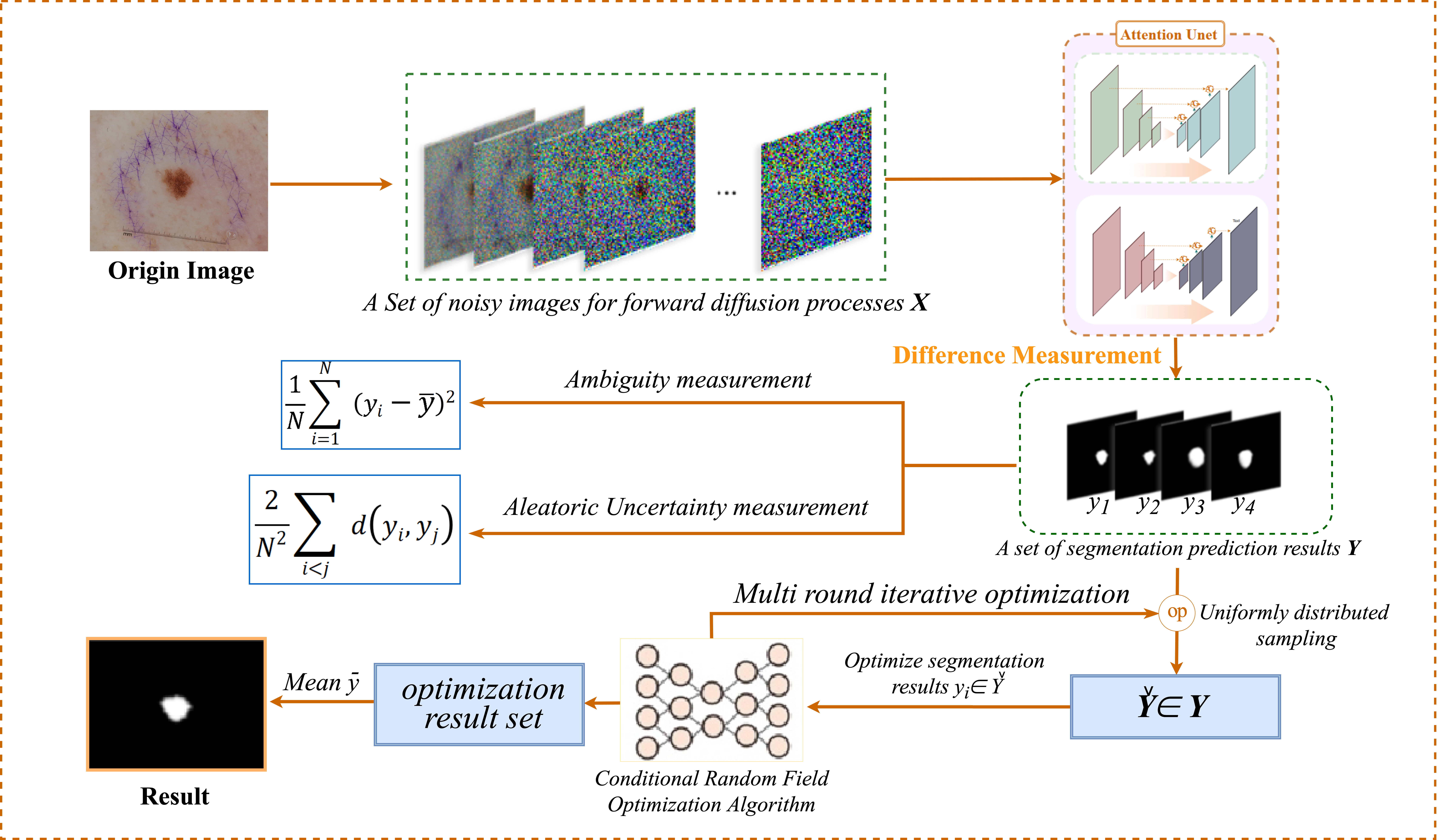}
    \caption{Strategies for measuring ambiguity and uncertainty in multi-output results and optimizing outcomes.}
    \label{fig:fig2}
\end{figure*}
\section{METHOD}
\subsection{Differentiation metrics}

We propose the following method:

Given a dataset X that includes annotations for two categories, where $c_0$ represents the "healthy" annotation and $c_1$ represents the "unhealthy" annotation, as illustrated in Figure (\ref{fig:fig1_long}). A trained diffusion model $f_\theta$ can perform weakly-supervised semantic segmentation using only these category annotations. By employing variational inference techniques \citep{hoDenoisingDiffusionProbabilistic2020c} to optimize the likelihood function, we can derive the loss function:

\begin{equation}
\label{eq:L}
    L=E_{(x,c)}E_{\overline{\alpha}_t}\|f_\theta(\sqrt{\bar{\alpha}_t}x_0+\sqrt{1-\bar{\alpha}_t}\varepsilon,c,\bar{\alpha}_t)-\varepsilon\|
\end{equation}

The Gaussian noise variance scales are set as $\{\beta_t\in\left(0,1\right)\}_{t=1}^\mathrm{T}$, with ${\bar{\alpha}}_t=\prod_{i=1}^{t}{1-\beta_i}$. Equation (\ref{eq:L}) is equivalent to maximizing the variational lower bound of the likelihood function \citep{hoDenoisingDiffusionProbabilistic2020c}.

Specifically, since the diffusion model $\mathrm{f_\theta(x_0,c,\overline{\alpha}_t)}$ is conditioned on the category label embeddings $c$, switching the category label embeddings $c$ during the generation process can steer the original image towards different image categories, thereby resulting in different output noises. The pixel-level differences between these two noises contain semantic information:

\begin{equation}
\label{eq:y}
    y=Bin(|f_\theta(x_t,c_0,\bar{\alpha}_t)-f_\theta(x_t,c_1,\bar{\alpha}_t)|,\delta)
\end{equation}

where $\mathrm{Bin}(\cdot,\delta)$ denotes the process of binarizing the model's output by setting a threshold $\theta$. By measuring the differences in noise levels between $c_0$ and $c_1$ and binarizing the difference results, the segmentation outcome $y$ corresponding to the location of malignant lesions for the categories can be obtained.

\subsection{Aleatoric Uncertainty Metrics}
To address the aleatoric uncertainties in the segmentation results, we quantify them from two perspectives: size of ambiguity areas and value of aleatoric uncertainty.

Aiming for this objective, we first leverage the characteristics of the diffusion model by sampling at different time steps n, achieving multiple output segmentation results, denoted as $Y\in\{y_1,y_2,...,y_t,...,y_n\}$.

Subsequently, we use the mean and variance as mathematical methods to represent the consistent and ambiguous areas within the segmentation results, respectively. The mean of the set of segmentation results represents the model's average prediction for a given input image, as it reflects the most consistent part across all model outputs and can be viewed as the “consistent area” of model predictions. The variance measures the deviation of each segmentation result from the mean, indicating the variability among predictions. Therefore, a larger variance suggests greater differences in model predictions within that area, indicating stronger ambiguity. These areas may require additional attention in medical diagnostics, as the model's predictive confidence in these regions is lower, and different experts may have varying interpretations during image segmentation.

The specific formulas for mean and variance are as follows:

\begin{equation}
\label{eq:yc}
    Y_{Coherence}=\overline{y}
\end{equation}

\begin{equation}
\label{eq:ya}
    Y_{Ambiguity}=\frac1N\sum_{i=1}^N(y_i-\bar{y})^2
\end{equation}

in which, $y_i$ represents the segmentation outcomes at different noise addition levels.

Next, utilizing the concept of the Generalized Energy Distance (GED), we quantify the uncertainty of outputs from the result set, expressed by the formula:

\begin{equation}
\label{eq:DGED}
    D_{GED}^2=\frac2{N^2}\sum_{i<j}d\left(y_i,y_j\right)-\frac1{N^2}\sum_{i=1}^Nd\left(y_i,y_i\right)-\frac1{N^2}\sum_{j=1}^Nd\left(y_j,y_j\right)
\end{equation}

where $\mathrm{d}(\cdot)$ is a distance function that represents the similarity between two segmentation results, and $y_i$,$y_j\in Y$. To intuitively measure the pixel-level prediction differences from the true annotations, we opt for the Euclidean distance for computation. The formula can be articulated as:

\begin{equation}
\label{eq:dpq}
    d(P,Q)=\sqrt{\sum_{i=1}^{w\times h}(p_i-q_i)^2}
\end{equation}

Herein, $p_i$ and $q_i$ respectively denote the pixel points in two segmentation results. Since the simple Euclidean distance measure is influenced by the image size, the formula introduces a factor of the image size to normalize the distance.

\begin{equation}
    d(P,Q)=\sqrt{\frac{\sum_{i=1}^{w\times h}{(p_i-q_i)^2}}{w\times h}}
\end{equation}

Through the calculations described above, we can obtain the Generalized Energy Distance $D_{GED}^2\in[0,\frac{\sqrt{2}}2]$, which serves as a metric for quantifying the differences between two probability distributions. It can be used to measure the distance between different prediction results within a set of segmentation outcomes. By calculating the GED, we can more accurately assess the uncertainty of the segmentation results and the diversity of model predictions. Larger values of GED indicate greater differences between segmentation results, whereas smaller values suggest more consistency among the results, further eliminating the interference caused by noise external to the target.

\subsection{Segmentation results optimization}

In the field of image segmentation, the Conditional Random Field (CRF) algorithm performs outstandingly \citep{vemulapalliGaussianConditionalRandom2016}, and it is often used in the post-processing stage of MIS to refine and optimize preliminary segmentation results. Therefore, for multiple output results, we adopt an improved version of the DenseCRF to optimize the results. DenseCRF considers the relationship between each pixel in the image and all other pixels, enabling it to capture finer details and a broader context. This allows it to provide more precise segmentation results in some cases.

The optimization process is as follows:

First, we initialize the set of segmentation results $Y$, where initially, the probability of each element being selected is set to an equal probability of $1/n$.

Second, the set of segmentation results undergoes $K$ iterations. In each iteration, a subset $\check{Y}$ is randomly sampled from $Y$ with a given probability. This step introduces randomness to explore different combinations of segmentation results in $Y$, thereby discovering and utilizing the complementary information among these results.

Third, DenseCRF is applied to each segmentation result in the sampled subset $\check{Y}$ to filter out noise and improve segmentation quality.

The core of DenseCRF is an energy function $E(\mathbf{x})$, as shown in formula 12. This function consists of two parts: a data term (Unary Potential), which can be directly obtained from the noise difference calculated by the formula (\ref{eq:dpq}) indicating the initial confidence of each pixel belonging to each category; and a smoothness term (Pairwise Potential), which depends on the relationship between pairs of pixels. This is usually defined using the intensity or position information of pixels in the original image, aiming to assign similar pixels to the same category while maintaining the clarity of edges. In this experiment, we design an appearance kernel based on the principle that nearby pixels with similar colors are likely to belong to the same class. The smoothness kernel is used to remove small isolated areas \citep{shottonTextonBoostImageUnderstanding2009}.

\begin{equation}
    E(\mathbf{x})=\sum_i\psi_u(x_i)+\sum_{i<j}\psi_p(x_i,x_j)
\end{equation}

\begin{equation}
    \psi_u(x_i)=-\log P(x_i)
\end{equation}

\begin{align}
    \psi_p(x_i,x_j) &= \mu(x_i,x_j)\left[w_1\underbrace{\exp\left(-\frac{\left|p_i-p_j\right|^2}{2\theta_\alpha^2}-\frac{\left|l_i-l_j\right|^2}{2\theta_\beta^2}\right)}_{\text{appearance kernel}} \right. \nonumber \\
    &\quad \left. + w_2\underbrace{\exp\left(-\frac{\left|p_i-p_j\right|^2}{2\theta_\gamma^2}\right)}_{\text{smoothness kernel}}\right]
\end{align}
in which, $P\left(x_i\right)$ denotes the probability that pixel $i$ belongs to label $x_i$, $\mu\left(x_i,x_j\right)$ is a penalty term for label inconsistency, which is 1 when $x_i\neq x_j$ and 0 otherwise. $p_i$ and $p_j$ represent the position coordinates of pixels $i$ and $j$ respectively,  $I_i$ and $I_j$ represent the corresponding color intensity or feature vectors, $w_1, w_2, \theta_\alpha, \theta_\beta \text {, and } \theta_\gamma$ model parameters used to adjust the influence of the Gaussian kernel.

The objective of DenseCRF is to achieve energy minimization, a process accomplished through the mean field algorithm. In this process, the category label of each pixel is adjusted in every iteration to reduce the overall energy.

\begin{equation}
    E^{(t+1)}(\mathbf{x})=\frac1{z_i}\mathrm{exp}\left(-\psi_u(x_i)-\sum_{j\neq i}\sum_{x_j}E^{(t)}(\mathbf{x})\psi_p(x_i,x_j)\right)
\end{equation}

In this formulation, $E^{(t)}\left(\mathbf{x}\right)$ represents the probability distribution of the label $x_i$ for pixel $i $ at iteration step $ t$, where $Z_i$ is the normalization constant that ensures $E(\mathbf{x})$ is a probability distribution. This step is iterated until $E(\mathbf{x})$ converges or the maximum number of iterations is reached, marking the end of DenseCRF optimization. 

Subsequently, the optimized segmentation results are averaged and then binarized to obtain the average prediction ${\bar{y}}_k$ for that iteration step. This step combines multiple optimized results to reduce bias and interference in individual predictions.

After completing all $K$ iterations, further averaging is performed on the average predictions ${\bar{y}}_k$ obtained from all iteration steps, with the calculation formula being:

\begin{equation}
    Y_{final}=\frac1K\sum_{k=1}^K\bar{y}_k
\end{equation}

Through these operations, we can synthesize information obtained across all iterations, smooth out the segmentation results, and reduce noise and errors, thereby achieving a more accurate and stable final segmentation prediction.

This method allows us to utilize the information about the uncertainty in model predictions to iteratively refine and improve the segmentation results.


\section{EXPERIMENTS}
\subsection{Model Training}
In this experiment, we utilize the ISIC2018 challenge dataset \citep{codellaSkinLesionAnalysis2018b}, \citep{tschandlHAM10000DatasetLarge2018}, which contains a large number of high-resolution images of skin lesions and their corresponding annotation information. After collecting the data, we conduct thorough data organization, which included 2500 images in the training set, 100 in the validation set, and 1000 in the test set. Furthermore, we use the annotations corresponding to the training set as masks to remove the lesion areas in the disease images, considering them as healthy skin, to aid the model in learning and acquiring features of healthy images.

To enhance the model's training effectiveness, we improve the data input method by adding various preprocessing techniques such as blurring, rotation, and sharpening to the input data.

During the model training process, based on Python 3.9.13 and PyTorch 2.0.1 frameworks, and conducted on an RTX 4070, we perform multiple experiments to identify well-performing hyperparameters. These included an image size of $128\times128$, a noise addition count $t=100$, a batch size of 4, a learning rate of 0.0005, and a training duration of 500 epochs. We train two diffusion models based on U-Net, with a total of 13.8M parameters for each training session, averaging a duration of 7 minutes and 11 seconds.

In the model testing phase, for the multiple outputs of the diffusion model, we set 10 time steps between 60 and 150, at intervals of 10, to quantify the consistency areas of segmentation results, ambiguity areas, calculate the GED values, and the subsequent optimization processes.

During the optimization process, we set three iterations, each time selecting 4 results from the outputs of 10 time steps for optimization calculations. Finally, we perform a second averaging calculation on the optimized results after three iterations.

\subsection{Index Evaluation}

The evaluation indicators we use in this experiment include Dice Coefficient, Jaccard Index, Precision and Recall. The calculation formulas are as follows:

\begin{subequations}
    \begin{align}
        &\text{Dice}=\frac{2\times|X\cap Y|}{|X|+|Y|} \\
        &\text{Jaccard}=\frac{|X\cap Y|}{|X\cup Y|} \\
        &\text{Precision}=\frac{TP}{TP+FP} \\
        &\text{Recall}=\frac{TP}{TP+FN}
    \end{align}
\end{subequations}

Herein, $X$ represents the segmentation results predicted by the model, while $Y$ represents the actual segmentation. Positive cases typically refer to the target areas that the model needs to identify and segment, namely, the lesion areas. Conversely, negative cases refer to areas other than the positive ones, namely, the non-target areas. 

These metrics are greatly beneficial in assessing the model's performance. The value of the Dice coefficient ranges from 0 to 1, with higher values indicating greater similarity between the predicted segmentation and the actual segmentation, that is, better model performance. The Jaccard index also has values ranging from 0 to 1, with higher values indicating a higher degree of overlap between the predicted and actual segmentations, meaning better model performance. High precision means fewer false positives, that is, the model is more accurate in marking positive cases. A high recall rate implies that the model can capture more actual positive cases, thereby reducing missed detections.

\section{RESULTS \& DISCUSSION}
\subsection{Medical Image Segmentation}

\begin{figure}[ht]
    \centering
    \includegraphics[width=\columnwidth]{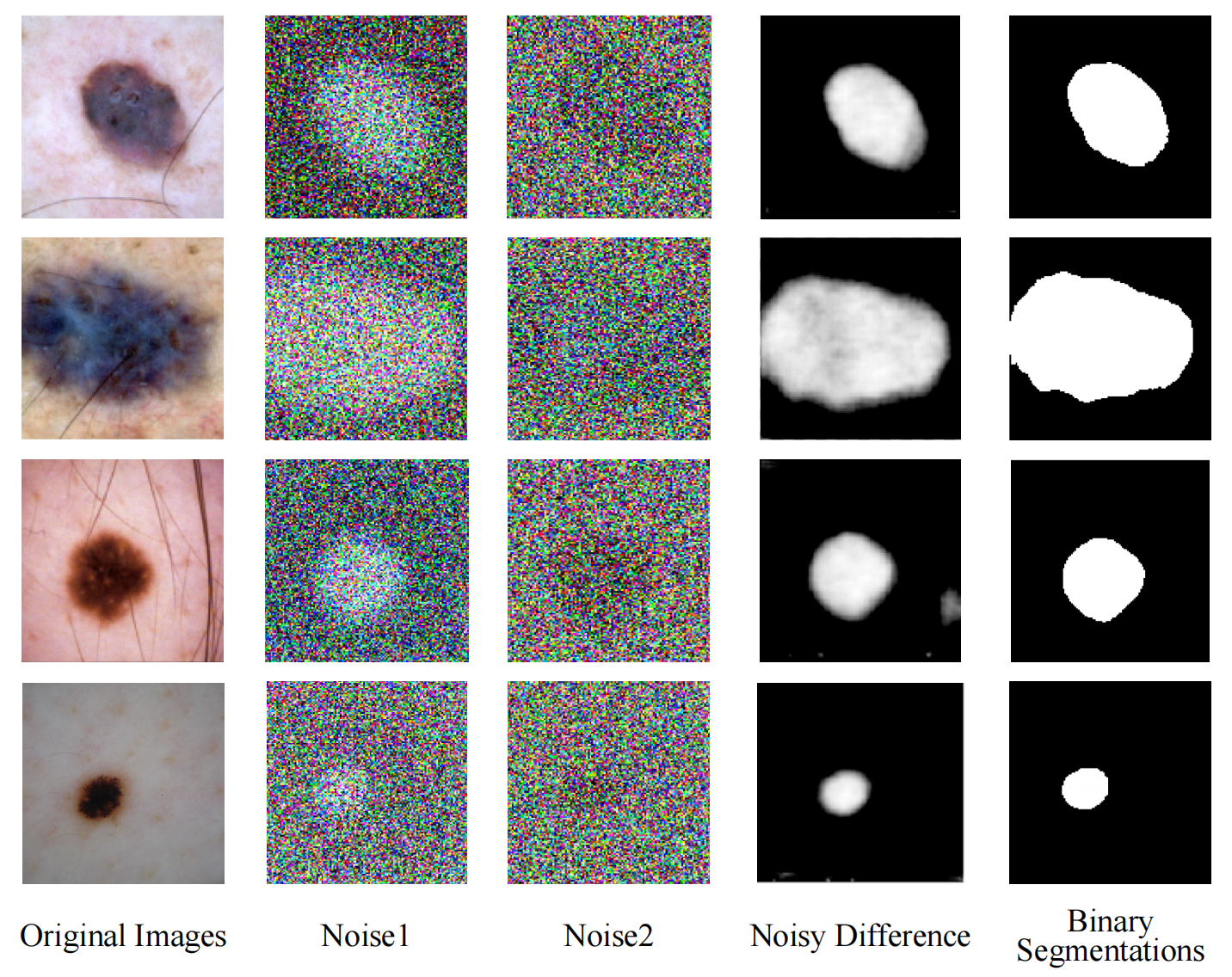}
    \caption{The experimental process includes the original images, noise under different semantics, noise difference grayscale images, and binarized segmentations.}
    \label{fig:process}
\end{figure}

The figure below showcases several distinct examples from our experimental process. (see Fig. \ref{fig:process}) It is observed that our experimental model can efficiently differentiate between skin and other interfering factors, such as hair and bubbles, which are disturbances during the image acquisition process. In the process of denoising with the diffusion model, it can treat interfering factors as noise and separate them from the skin, thereby focusing more on the differences between lesions and normal skin when measuring noise diversity. This approach results in image segmentation outcomes that are closer to the ground truth.

Particularly in the areas where skin meets hair and at the lesion borders, our model demonstrated superior performance compared to the traditional U-Net model. This strategy enhances the model's robustness and improves its capability to capture details, which is especially important when dealing with highly complex and variable data such as medical images.

\subsection{Aleatoric Uncertainty Measurement}
After completing the segmentation task, we utilize the model's multi-output capability to generate multiple different output results at various time steps. The specific output results are illustrated in the figure below:

\begin{figure}[ht]
    \centering
    \includegraphics[width=0.8\columnwidth]{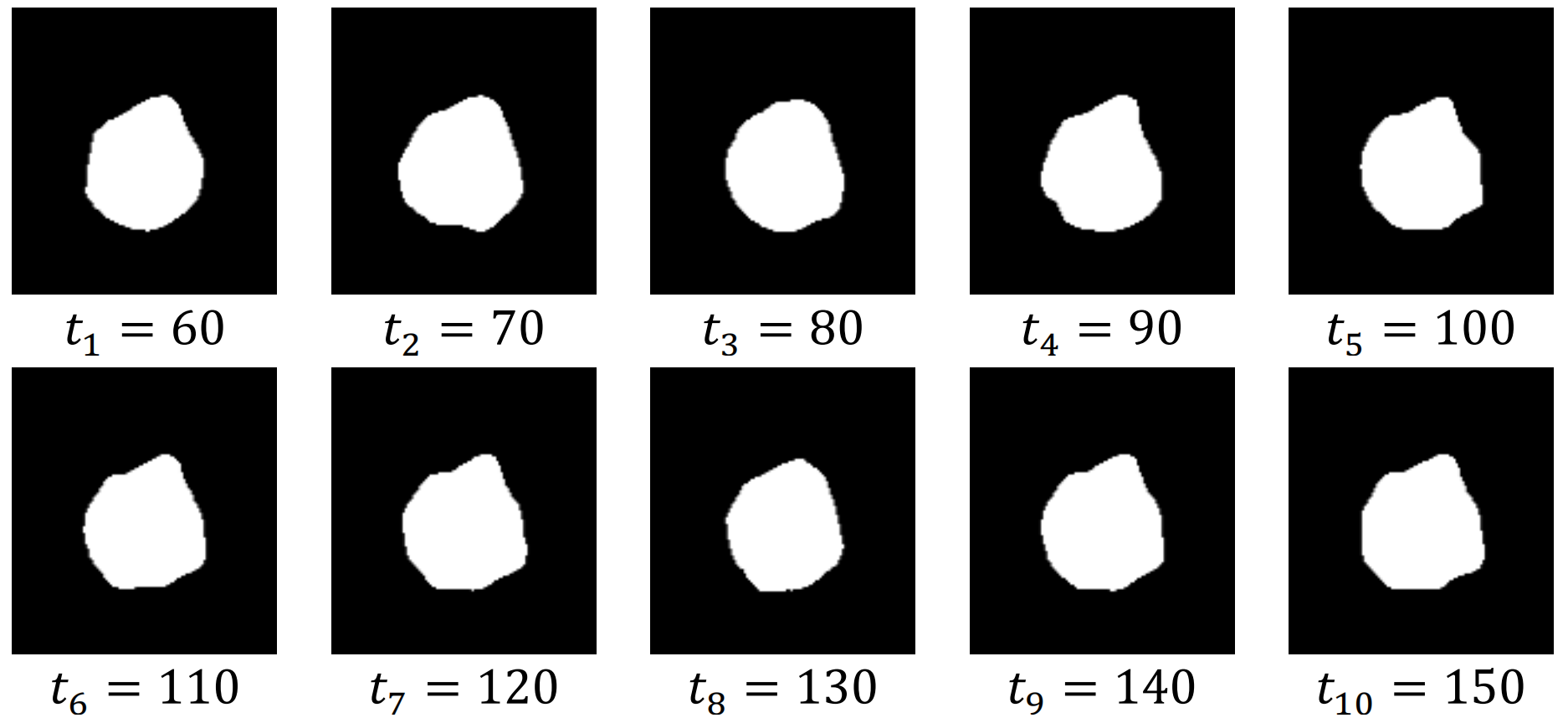}
    \caption{Segmentation results obtained under different noise addition iterations (from 60 to 150).}
    \label{fig:multi_results}
\end{figure}

By employing the formula (\ref{eq:yc}) and (\ref{eq:ya}), we calculate the mean and variance of the aforementioned ten output results. This allows us to identify the consistency and ambiguity regions of the segmentation results, which are specifically displayed in the following figure.

\begin{figure}[ht]
    \centering
    \includegraphics[width=0.6\columnwidth]{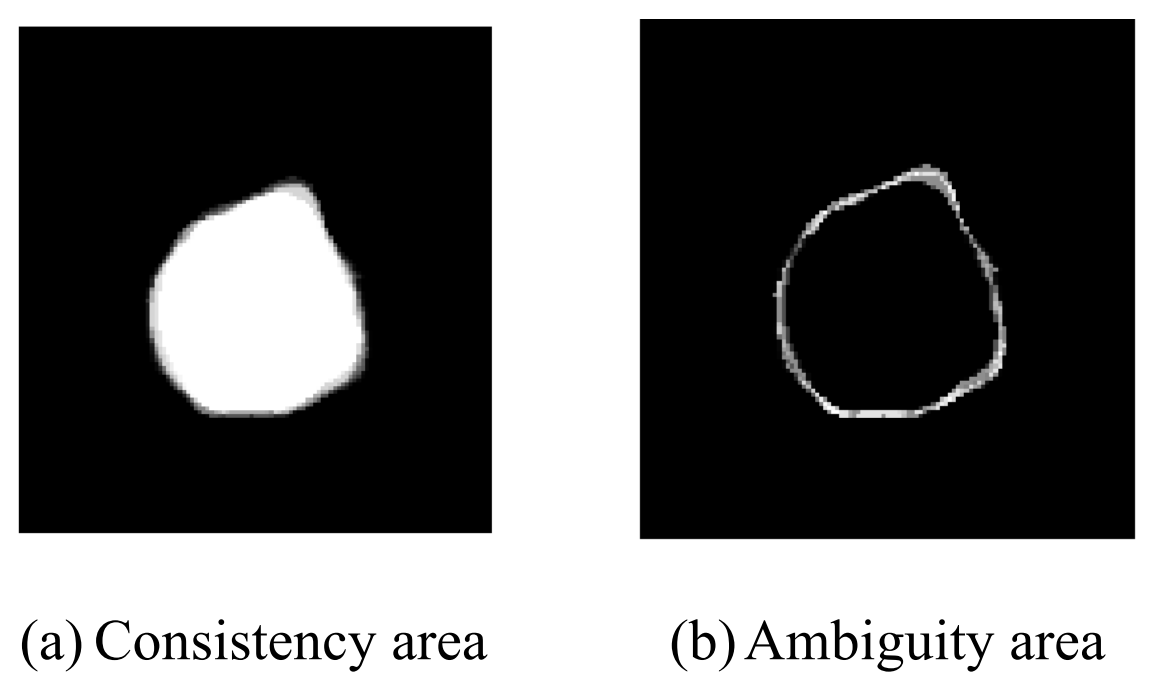}
    \caption{After calculation, the consistent and ambiguous regions are obtained.}
    \label{fig:ambiguous}
\end{figure}

Additionally, according to the measurement method provided by the formula (\ref{eq:DGED}), the uncertainty value of the segmentation result can be calculated to be 0.15. Through testing, on the entire test set, the average uncertainty of our segmentation results is 0.13.

These results allow doctors to visually observe the model's segmentation performance. In practical applications, doctors can use this information to focus on the ambiguous regions within the segmentation results, providing effective decision sup-port for clinical diagnosis and further research. Moreover, our approach to uncertainty measurement with a defined range yields numerical results within a specific range, offering clear and effective assistance for subsequent optimization research.

\subsection{Optimization for Multi-output}

By analyzing the multiple output results generated by the model, we employ the DenseCRF algorithm to optimize these results. The specific outcomes are illustrated in the figure below.

\begin{figure}[ht]
    \centering
    \includegraphics[width=0.2\columnwidth]{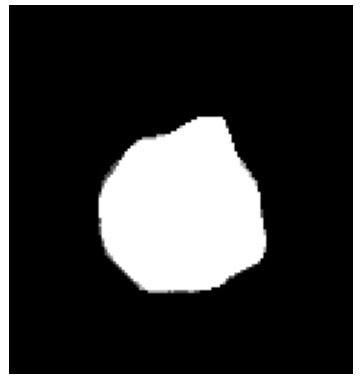}
    \caption{After calculation, the consistent and ambiguous regions are obtained.}
    \label{fig:final}
\end{figure}

Comparing the multiple output results in the figure \ref{fig:multi_results}, it is evident that through iterative optimization steps, we further smooth the segmentation results, reducing noise and errors to achieve more accurate and stable final segmentation predictions. This process not only enhanced the precision of the segmentation results but also made full use of the uncertainty information from the model's predictions, offering a new avenue for precise medical image analysis.

After foregoing processes, we obtain the following results, as detailed in Table 1. Additionally, we list the outcomes of some segmentation models based on U-Net on the ISIC 2018 dataset.

\begin{table}[!ht]
    \centering
    \caption{The comparison of the segmentation methods.}
    \begin{tabular}{ccccc}
    \toprule
        \textbf{Method} & \textbf{Dice} & \textbf{Precision} & \textbf{Recall} & \textbf{Jaccard} \\ 
        \midrule
        \textbf{U-Net} & 0.838 & 0.863 & 0.868 & 0.724 \\ 
        \textbf{DU-Net} & 0.852 & 0.894 & 0.867 & \textbf{0.749} \\ 
        \textbf{RU-Net} & 0.85 & 0.892 & 0.855 & 0.743 \\ 
        \textbf{AttnU-Net} & 0.86 & 0.861 & \textbf{0.884} & 0.746 \\ 
        \textbf{DiffSeg} & \textbf{0.864} & \textbf{0.897} & 0.882 & 0.735 \\
    \bottomrule
    \end{tabular}
    \label{The comparison}
\end{table}

From the table data, it is evident that compared to other U-Net-based models, our experimental model demonstrates superior performance in terms of the Dice coefficient and Precision.

\section{CONCLUSION}

In this study, we introduce a novel MIS model, known as DiffSeg, specifically designed to ad-dress two major challenges faced by traditional segmentation models in the context of processing highly complex medical images: One is the excessive dependency on large-scale annotated datasets; and the other is the difficulty in providing reliable uncertainty measurements. By introducing a denoising diffusion probability model, we skillfully leverage the model's capability to extract features (i.e., noise) from images with different semantics and utilize the differences in image noise to achieve precise semantic segmentation under weak supervision conditions. This approach significantly improves segmentation accuracy and demonstrates the practicality of generative models in the field of MIS. Experimental results show that the DiffSeg model outperforms existing U-Net-based segmentation methods on the ISIC 2018 dataset, particularly in terms of the Dice coefficient and recall rate.

Furthermore, we explore a measurement method for the aleatoric uncertainty of segmentation results by utilizing the model's multi-output capability to quantify the consistency and ambiguity regions of the segmentation results. We also define an uncertainty calculation method with a determined range using GED. Based on the multi-output results, we optimize the segmentation results using the DenseCRF algorithm. This strategy not only improves the accuracy of the segmentation results but also provides intuitive decision support for doctors, helping to reduce repetitive labor in practical applications.

In summary, the DiffSeg model proposed in this study offers a new research direction for the application of generative models in the field of medical image segmentation. Our future work will focus on further enhancing the model's generalization ability to achieve higher accuracy and practicality in a broader range of medical image segmentation tasks. We hope our research will inspire more researchers and promote the development of advanced medical image analysis technologies.

\section{ACKNOWLEDGMENTS}

Project 202310611112 supported by National Training Program of Innovation and Entrepreneurship for Undergraduates.

\bibliography{ecai-template/reference}

\end{document}